%% file: main.tex
\documentclass{article}

\usepackage{arxiv}

\usepackage[utf8]{inputenc} 
\usepackage[T1]{fontenc}    
\usepackage{hyperref}       
\usepackage{url}            
\usepackage{booktabs}       
\usepackage{amsfonts}       
\usepackage{amsmath}        
\usepackage{nicefrac}       
\usepackage{microtype}      
\usepackage{lipsum}		
\usepackage{graphicx}
\usepackage{float}
\usepackage{natbib}
\usepackage{doi}
\usepackage{setspace} 
\graphicspath{ {.} }

\title{A Model Generalization Study in Localizing Indoor Cows with COw LOcalization (COLO) dataset}

\author{
    \href{https://orcid.org/0009-0001-8932-142X}{\includegraphics[scale=0.06]{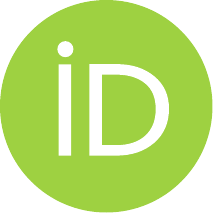}\hspace{1mm}Mautushi Das} \\
	School of Animal Sciences\\
	Virginia Tech\\
	Blacksburg, VA 24061 \\
	\texttt{mautushid@vt.edu} \\
	\And
    \href{https://orcid.org/0000-0002-8254-8090}{\includegraphics[scale=0.06]{orcid.pdf}\hspace{1mm}Gonzalo Ferreira} \\
	School of Animal Sciences\\
	Virginia Tech\\
	Blacksburg, VA 24061 \\
	\texttt{gonf@vt.edu} \\
	\And
	\href{https://orcid.org/0000-0002-2018-0702}{\includegraphics[scale=0.06]{orcid.pdf}\hspace{1mm}C. P. James Chen} \thanks{Corresponding author: James Chen \texttt{<niche@vt.edu>}}\\
	School of Animal Sciences\\
	Virginia Tech\\
	Blacksburg, VA 24061 \\
	\texttt{niche@vt.edu}
}

\renewcommand{\shorttitle}{An Model Generalization Study with COLO dataset}

\begin{document}
\maketitle

\setstretch{1.2}

\input{_0_abstract.tex}
\newpage
\input{_1_introduction.tex}
\input{_2_approach.tex}

\input{_3_results.tex}

\section*{Acknowledgments}

This research was supported by the USDA Hatch Research Project funding VA-160196. The authors acknowledge Advanced Research Computing at Virginia Tech for providing computational resources and technical support that have contributed to the results reported within this paper. URL: https://arc.vt.edu/. During the preparation of this work the author(s) used ChatGPT in order to ensure the grammar and clarity of the manuscript. After using this tool/service, the author(s) reviewed and edited the content as needed and take(s) full responsibility for the content of the publication.

\newpage

\bibliographystyle{plainnat}
\bibliography{references}

\newpage

\input{_9_appendix.tex}

\end{document}

%% file: _0_abstract.tex
\begin{abstract}

Precision livestock farming (PLF) increasingly relies on advanced object localization techniques to monitor livestock health and optimize resource management. This study investigates the generalization capabilities of YOLOv8 and YOLOv9 models for cow detection in indoor free-stall barn settings, focusing on varying training data characteristics such as view angles and lighting, and model complexities. Leveraging the newly released public dataset, COws LOcalization (COLO) dataset, we explore three key hypotheses: (1) Model generalization is equally influenced by changes in lighting conditions and camera angles; (2) Higher model complexity guarantees better generalization performance; (3) Fine-tuning with custom initial weights trained on relevant tasks always brings advantages to detection tasks. Our findings reveal considerable challenges in detecting cows in images taken from side views and underscore the importance of including diverse camera angles in building a detection model. Furthermore, our results emphasize that higher model complexity does not necessarily lead to better performance. The optimal model configuration heavily depends on the specific task and dataset, highlighting the need for careful model selection tailored to the particular application. Lastly, while fine-tuning with custom initial weights trained on relevant tasks offers advantages to detection tasks, simpler models do not benefit similarly from this approach. It is more efficient to train a simple model with pre-trained weights without relying on prior relevant information, which can require intensive labor efforts. Future work should focus on adaptive methods and advanced data augmentation to improve generalization and robustness. This study provides practical guidelines for PLF researchers on deploying computer vision models from existing studies, highlights generalization issues, and contributes the COLO dataset containing 1254 images and 11818 cow instances for further research.

\end{abstract}
\keywords{Object detection \and Cows \and Model generalization \and Model selection}

%% file: _1_introduction.tex
\section{Introduction}

\subsection*{Object Localization and Its Applications}

Localizing livestock individuals from images or videos has become an essential task in precision livestock farming (PLF) \cite{fernandes2020image}. Such techniques allow farm operators to manage animal well-being and health in real-time, optimizing their resource management and improving sustainability \cite{morrone2022industry, hao2023cattle}. Technically speaking, in the field of computer vision (CV), which is a subfield of artificial intelligence (AI) that focuses on translating visual information into actionable insights, localization tasks can be further categorized into object detection, object segmentation, and pose estimation. Object detection is the simplest form among these tasks, localizing objects of interest by enclosing them within a rectangular bounding box defined by x and y coordinates, pixel width, and pixel height \cite{viola2001rapid}. Successful instances in this category include YOLO (You Only Look Once) \cite{redmon2016you}, Faster R-CNN (Region Convolutional Neural Networks) \cite{girshick2015fast}, and SSD (Single Shot MultiBox Detector) \cite{liu2016ssd}. These models have been adopted and applied by animal scientists for detection in precision livestock farming. For example, a study \cite{yu2022automatic} leveraged the DRN-YOLO model \cite{xu2020improved} to predict the eating behavior of dairy cows. This approach automates the assessment of feeding behavior, a critical indicator of cow health and productivity, and has saved labor efforts in complex farm settings. Another notable work is presented in \cite{nasirahmadi2019deep}, where the authors developed a posture detection system for pigs using deep learning models such as Faster R-CNN, SSD, and R-FCN, coupled with 2D imaging. This system accurately identifies standing and lying postures of pigs under commercial farm settings.

To achieve finer localization, object segmentation is employed to outline object contours pixel-wise, while pose estimation is performed by orienting and marking the key points of the object \cite{hariharan2015hypercolumns}. Some popular object segmentation models include Mask R-CNN \cite{he2017mask}, MS R-CNN \cite{huang2019mask}, and U-Net \cite{siddique2021u}. This method of segmentation has also been applied in the field of PLF. In the study \cite{noe2022automatic}, the authors developed a method using Mask R-CNN \cite{he2017mask} to segment and outline cattle in feedlots. Their technique enhances images and extracts key frames to accurately detect cattle, achieving superior precision with a mean pixel accuracy of 0.92. This supports advanced, real-time monitoring of cattle in PLF. Another study group \cite{tu2021automatic} developed the PigMS R-CNN framework \cite{huang2019mask} to enhance the monitoring of group-housed pigs. This framework employs a 101-layer residual network along with a feature pyramid network and soft non-maximum suppression to effectively detect and segment pigs, thereby improving the accuracy of identifying and locating individual pigs in complex environments.

\subsection*{Model Generalization, Pre-Training, and Fine-Tuning}

Although implementing image-based systems in livestock production has become more common, current studies primarily focus on accuracy in homogenous environments and rarely address the challenges of model generalization. How a model can generalize to new environments is critical when farm operators deploy existing CV models in their own settings. Good generalization performance ensures that the model can reproduce similar results as reported in the original study, even in new environments with different conditions. Factors such as camera angles and the presence of occlusions can impact generalization in the deployment environment. Deploying the same model in a new environment does not necessarily guarantee the same performance as reported in the original study. Li et al. \cite{li2021practices} also pointed out that the lighting conditions on farms in real applications can be highly variable, leading to poor generalization performance.

One explanation for poor generalization is the discrepancy between the pre-training process and the specific use case. Most CV models are released with pre-trained weights, obtained from training on a large-scale dataset. For example, the COCO dataset \cite{lin2014microsoft} is a general-purpose dataset containing over 200,000 images and a wide range of object categories, such as vehicles and household items. Directly deploying a model pre-trained on the COCO dataset to detect cows in a farm setting may not ensure satisfactory performance, as the dataset does not contain enough cow instances in different view angles or occlusions. To alleviate this discrepancy, fine-tuning is a common practice that modifies the prediction head of the pre-trained model and updates the weights on a new dataset more relevant to the specific use case. Most application studies have adopted this approach to improve model generalization on their specific tasks \cite{han2021pre,guirguis2022cfa,gupta2023novel}.

Nevertheless, fine-tuning is not guaranteed to be successful, as the outcome depends on both the quantity and quality of the annotated dataset. For example, Zin et al. \cite{zin_automatic_2020} deployed an object detection model to recognize cow ear tags in a dairy farm. Although the model achieved a high accuracy of 92.5\% in recognizing the digits on the ear tags, more than 10,000 images were required for fine-tuning. Assembling such a large dataset is labor-intensive and requires specific training in annotating the images. The annotated dataset is rigorously organized in a specific format. For example, the COCO annotation format \cite{lin2014microsoft} stores image information, object class, and annotations of the entire dataset in one nested JSON format. In contrast, the YOLO format \cite{ultralytics2023datasets}, another common format for object localization, stores information of one image in one text file, with each line representing one object instance in the image. Additionally, unlike the COCO format that stores bounding box coordinates in absolute pixel values, the YOLO format stores the coordinates in relative values to the image size. These technical details are key to valid annotations, which are usually facilitated by professional annotation tools such as Labelme \cite{labelme2023}, CVAT \cite{cvat2023}, or Roboflow \cite{roboflow2023}.

\subsection*{Model Complexity and Performance}

Another factor affecting model generalization is model complexity. Generally, model complexity is quantified by the number of learnable parameters in a model \cite{hu2021model}. A more complex model can often generalize better to unseen data with high accuracy. However, this high complexity also comes at the cost of computational resources in the form of memory or time \cite{justus2018predicting}. The computational cost may further limit how models can be deployed in real-world applications, where real-time processing or edge computing is desired for fast or compact systems. For instance, the VGG-16 model \cite{simonyan2014very} has 138 million parameters and requires a video memory of at least 8GB, while the ResNet-152 \cite{he2016deep} has around 60 million parameters with a recommended video memory of 11GB. Recent models for object detection, such as YOLOv8 \cite{ultralyticsYOLOv8} and YOLOv9 \cite{wang2024yolov9}, have been developed in different sizes, providing a flexible choice for researchers to balance between generalization performance and computational cost. In YOLOv8, the spectrum of model complexity ranges from the highly intricate YOLOv8x, containing 68.2 million parameters, to more streamlined variants like YOLOv8n with only 3.2 million parameters. The memory demand for the model architecture alone, without considering the intermediate results during training or inference, is larger by a factor of 21 for YOLOv8x (136.9 megabytes) compared to YOLOv8n (6.5 megabytes). Therefore, the trade-off between model complexity and computational cost is a critical factor to consider when deploying CV models in real-world scenarios.

\subsection*{YOLO Models}

Before YOLO, object detection methods typically involved either using “sliding windows with classifier” or “region proposals with classifier.” The sliding windows method required running the classifier hundreds or thousands of times per image. On the other hand, advanced region proposal-based approaches divided the task into two steps: first, identifying potential object regions (i.e., region proposals) and then applying a classifier to these regions. In contrast, YOLO models are capable of performing object detection in a single pass through the network, which is why the acronym YOLO stands for “You Only Look Once.”

YOLOv8 \cite{ultralyticsYOLOv8}, building on the YOLOv5 \cite{Jocher2020YOLOv5} architecture, incorporates the C2F module (cross-stage partial bottleneck with two convolutions), a refinement of the CSPLayer of YOLOv5 featuring two convolutional operations. It employs SiLU activation over traditional ReLU and Sigmoid \cite{elfwing2018sigmoid} for smoother gradient flow, enhancing CNN performance. The module divides input from a convolutional layer, processes one half through bottleneck layers (offering two types: with and without shortcuts similar to ResNet \cite{targ2016resnet}), then merges it back for further convolution. This design, along with a spatial pyramid pooling fast (SPPF) layer in its backbone, supports efficient feature pooling and multi-scale detection by using three distinct heads, thereby optimizing object detection across varying sizes. Furthermore, YOLOv8 innovates with an anchor-free approach, directly predicting bounding boxes and confidence scores, thus simplifying the network and reducing computational overhead \cite{law2018cornernet,duan2019centernet,tian2019fcos}.

Deep learning models, including the YOLO family, encounter an information bottleneck issue \cite{tishby2015deep,tishby2000information}, where the retention of input information diminishes as data is compressed into features. This loss is exacerbated in deeper network layers, often leading to reduced model efficacy. One approach to mitigate this involves expanding the model’s width, i.e., increasing the number of parameters, which allows for broader feature mapping and potentially recaptures lost information. However, simply increasing model size can lead to unreliable data outputs and does not proportionally enhance model performance.

YOLOv9 addresses these challenges through innovations like Programmable Gradient Information (PGI) and the Generalized Efficient Layer Aggregation Network (GELAN) \cite{wang2024yolov9}. PGI optimizes gradient generation to minimize deep layer information loss, featuring a main branch for inference and auxiliary branches for enhanced training. GELAN, by integrating and pooling convolutional layers, ensures robust feature retention. This adaptive system notably boosts inference speed by 20\% \cite{wang2024yolov9} on the COCO dataset \cite{lin2014microsoft}, while its multi-level auxiliary information facilitates the detection of objects across varying sizes, making YOLOv9 particularly effective in identifying smaller objects compared to its predecessors.

\subsection*{Public Datasets}

A public dataset helps the community to develop methodology based on the same baseline. One famous example in computer vision is the ImageNet dataset \cite{deng2009imagenet}, which serves as a benchmark for image classification. AlexNet \cite{krizhevsky2017imagenet}, the winner of the ImageNet Large Scale Visual Recognition Challenge in 2012, demonstrated its outstanding capability to classify images in the ImageNet dataset using Rectified Linear Units (ReLU) as the activation function, rather than the traditional sigmoid function. The success of AlexNet accelerated the development of CV models in the following years, such as VGG \cite{karen2014very}, GoogLeNet \cite{szegedy2015going}, ResNet \cite{targ2016resnet}, and DenseNet \cite{huang2017densely}. However, similar to the challenges that pre-trained models face in specific use cases, a generic public dataset, such as ImageNet \cite{deng2009imagenet} and COCO \cite{lin2014microsoft}, may not be sufficient for PLF applications. 

There have been efforts to create public datasets for livestock scenarios. For example, the CattleEyeView dataset was collected to support applications like cattle pose estimation and behavior analysis, providing extensive annotations across 30,703 frames from top-down video sequences of cows \cite{ong2023cattleeyeview}. Another study \cite{t2020long} leverages a public dataset for pigs comprising 3600 images from 12 videos of group-housed pigs. The dataset is particularly designed for applications such as pig tracking. Additionally, the "OpenCows2020" dataset, developed by researchers from the University of Bristol, is a public dataset specifically designed for advancing non-intrusive monitoring of cattle. It supports precision farming applications such as automated productivity assessment, health and welfare monitoring, and veterinary research, including behavioral analysis and disease outbreak tracing. The dataset consists of 11,779 images with 13,026 labeled objects, mainly focusing on cattle \cite{visualization-tools-for-opencows2020-dataset}. 

\subsection*{Study Objectives}

This study aims to use YOLO-family models to explore model generalization across varying environmental settings and model complexities within the context of indoor cow localization. Object detection, being the simplest form of localization, serves as an ideal baseline for extending to more complex tasks such as segmentation and posture estimation. Starting with object detection allows for a clear and foundational understanding of model behavior and performance, which can then inform and enhance the approach to more complex tasks. Consequently, this study examines three hypotheses:

\begin{itemize}
    \item \textbf{Model generalization is equally influenced by changes in lighting conditions and camera angles.} Should camera angles prove more impactful than lighting conditions, it would be advisable to prioritize camera placement when deploying CV models in new environments.
    \item \textbf{Higher model complexity guarantees better generalization performance.} If a highly complex model does not ensure superior performance, future studies might consider adopting less computationally demanding models that still enhance performance.
    \item \textbf{The advantages of using fine-tuned models as initial training weights are persistent over pre-trained models.} If the advantages diminish as the training sample size increases in a similar cow localization task but different environments, the fine-tuning efforts may be deemed unnecessary when the deployment environment varies over multiple locations on a farm.
\end{itemize}

To facilitate these investigations, a public dataset named COws LOcalization (COLO) \cite{COLODataset2023} will be developed and made available to the community. The findings of this study are expected to provide practical guidelines for Precision Livestock Farming (PLF) researchers on deploying CV models, considering available resources and anticipated performance.

%% file: _2_approach.tex
\section{Materials and Methods}

\subsection*{Cow Husbandry}

All procedures involving cow handling and image capturing were conducted in accordance with ethical guidelines and approved by the Virginia Tech Institutional Animal Care and Use Committee (IACUC \#22-146). The cows studied were part of the dairy herd at the Virginia Tech Dairy Complex in Blacksburg, Virginia, USA, which comprises approximately ~80\% Holstein and ~20\% Jersey cows. For the 'External' setting, the study included  100\% Holstein cows. The milking cows were housed in pens within a free-stall barn, featuring two rows of sand-bedded stalls, headlocks at the feed bunk, and two water troughs per pen. The stocking density was maintained at 100\% (i.e., one cow per stall). Heat stress was managed using automatic 48-inch diameter fans positioned over the stalls and feeding alleys. Cows were milked twice daily at 1:00 am and 12:00 pm in a double-twelve parallel milking parlor. They were fed ad libitum (with less than 5\% refusals) once daily at 8:00 am with a total mixed ration (TMR) consisting of approximately 42\% corn silage, 8\% grass hay, and 50\% concentrate on a dry matter basis. Manure from the stalls was removed at each milking session by personnel driving the cows to milking. Manure from the walking alleys within the pen was cleared two or three times daily using an automatic flushing system with recycled water. Fresh or recycled sand was added on a weekly basis.

\subsection*{Image Dataset}

The images in this study were collected using the Amazon Ring camera model Spotlight Cam Battery Pro (Ring Inc.), which offers a real-time video feed of dairy cows. Three cameras were installed in the barn: two at a height of 3.25 meters (10.66 feet) above the ground covering an area of 33.04 square meters (355.67 square feet). One camera provided a top view while the other was angled approximately 40 degrees from the horizontal to offer a side view of the cows. These are hereafter referred to as \textit{the Top-View camera} and \textit{the Side-View camera}, respectively. A third camera, termed \textit{the external camera}, was set at a lower height of 2.74 meters (9.00 feet) and covered a larger area of 77.63 square meters (835.56 square feet). Positioned 10 degrees downward from the horizontal, it captured a challenging perspective prone to occlusions among cows.

\begin{figure}[h]
    \centering
    \includegraphics[width=1\textwidth]{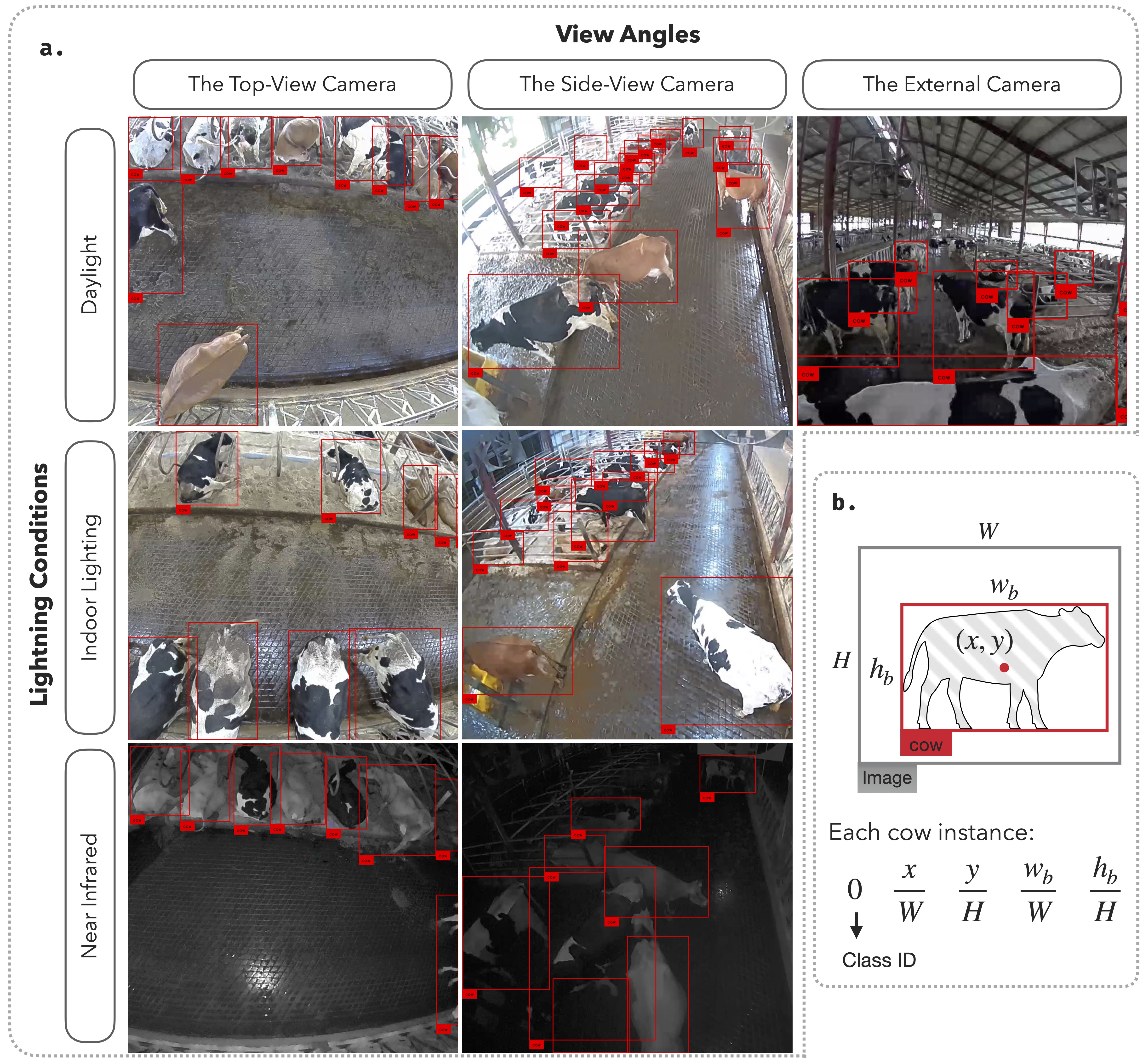}
    \caption{Overview of the COLO dataset. 1a. Seven instance images from the dataset with red bounding boxes labeling the location of cows. The columns show three different view angles: top-view, side-view, and external. The rows show three different lighting conditions: daylight, indoor, and near-infrared. 1b. An example of the annotated image in YOLO format. W, H, $w_b$, and $h_b$ represent the width, height, width of the bounding box, and height of the bounding box, respectively. x and y represent the center coordinates of the bounding box.} 
    \label{fig:dataset}
\end{figure}

Images were captured using an Ring Application Programming Interface (API) \citep{greif_dgreifring_2024}, configured to record a ten-second video clip every 30 minutes continuously for 14 days. Since the image quality relies on the camera's internet connection, which was occasionally unstable, some images were found to be tearing or unrecognizable. Hence, the resulting dataset was manually curated for consistent quality, comprising 504 images from \textit{the Top-View camera}, 500 from \textit{the Side-View camera}, and 250 from \textit{the external camera}. These images were further categorized based on the lighting conditions: for \textit{the Top-View camera}, 296 images were captured during daylight, 118 in the evening under artificial lighting, and 90 as near-infrared images without artificial light. From \textit{the Side-View camera}, 113 images were taken in the evening, and 97 as near-infrared images. All images from \textit{the external camera} were captured during the day. Image examples are shown in \textbf{Figure~\ref{fig:dataset}a}.

The image annotations were conducted using an online platform, Roboflow \cite{roboflow2023}, to define cow positions in the images. The bounding boxes were manually drawn to enclose the cow contours, providing the coordinates of the top-left corners and the width and height of the boxes. If cows were partially occluded, the invisible parts were inferred based on the adjacent visible parts. If the cow position was too far from the camera, making important body features such as the head, tail, and legs unrecognizable, the cow was excluded from the dataset. The final annotations were saved in the YOLO format \cite{ultralytics2023datasets}, where annotations were stored in a text file with one row per cow in the image, each row containing the cow's class, center coordinates, width, and height of the bounding box. The graphical representation of the annotated images is shown in \textbf{Figure~\ref{fig:dataset}b}.

\subsection*{Model Training}

The model training was implemented using the Python library Ultralytics \cite{ultralytics}. The model hyperparameters were set to the default values in the library. The training epochs were set to 100, and the batch size was set to 16. The implemented data augmentation included randomly changing the image color hue, saturation, and exposure to improve the model's generalization to different lighting conditions. Geometry augmentation was also applied by randomly flipping the images horizontally, copying and pasting to mix up object instances across multiple images to increase data diversity, and randomly scaling the images to simulate different distances between the camera and the cows. The details of the hyperparameters are shown in \textbf{Table~\ref{tab:hyperparameters}}. The training was conducted on an NVIDIA A100 GPU (NVIDIA, USA) with 80GB video memory provided by Advanced Research Computing at Virginia Tech.

\subsection*{Model Evaluation}

The examined YOLO models are object detection models that return positions of detected objects (i.e., cows in this study) for the evaluated images. The detections are represented by a list of bounding boxes. Regardless of specific procedures among YOLO variants for computational efficiency, such as YOLOv8, which integrates objectness scores and conditional class probabilities into a single confidence score, each detection generally consists of $4+c$ elements: the xy-coordinates, width, and height of the bounding box, and the $c$ confidence scores indicating the probability of the object belonging to each of the $c$ classes. The class with the highest confidence score is considered the predicted class of the object. To evaluate the model performance, two aspects are considered: the localization accuracy and the classification accuracy. The localization accuracy is measured by the Intersection over Union (IoU) between the predicted bounding box and the ground truth bounding box. On the other hand, the classification accuracy is measured by the precision and recall given the confidence threshold. If the confidence score of a detection is higher than the threshold, the detection is considered a positive detection. Otherwise, the detection is neglected. Combining the localization and classification accuracy, the mean Average Precision ($mAP$) averages the area under the precision-recall curve across all the classes. The curve is generated by varying the confidence threshold from 0 to 1 given an IoU threshold. In this study, four metrics were used in the evaluation: the precision and recall at the confidence threshold of 0.25 and IoU threshold of 0.5, the mAP at the IoU threshold of 0.5 (noted as $\text{mAP@{0.5}}$), and the averaged mAP at varying IoU thresholds ranging from 0.5 to 0.95 (noted as $\text{mAP@{0.5:0.95}}$).

\subsection*{Study 1: Benchmarking Model Generalization Across Different Environmental Conditions}

\begin{figure}[h]
    \centering
    \includegraphics[width=0.8\textwidth]{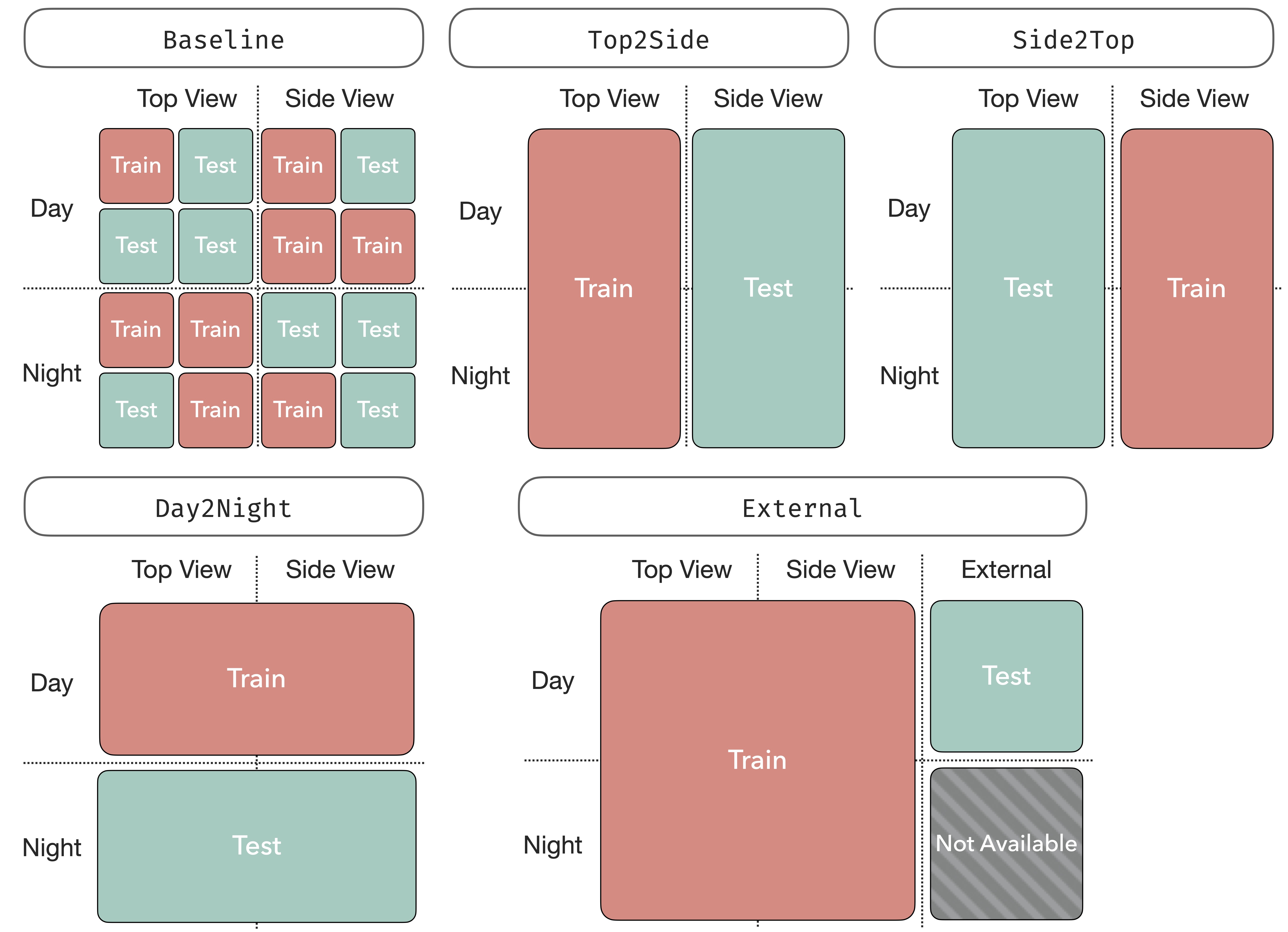}
    \caption{Cross-validation configurations. The training and testing sets were split into five different configurations: Baseline, Top2Side, Side2Top, Day2Night, and External.}
    \label{fig:splits}
\end{figure}

To compare the performance drop between different view angles and lighting conditions, we designed a cross-validation strategy where models were trained on one dataset configuration and tested on another. There are five training configurations in this study (\textbf{Figure~\ref{fig:splits}}):

\begin{itemize}
    \item \textbf{Baseline:} The model was trained and evaluated on the dataset characterized for all conditions, including top-view, side-view, daylight, evening, and near-infrared images. The images did not overlap between the training and evaluation sets.
    \item \textbf{Top2Side:} The model was trained on the top-view images and evaluated on the side-view images.
    \item \textbf{Side2Top:} The model was trained on the side-view images and evaluated on the top-view images.
    \item \textbf{Day2Night:} The model was trained on the daylight images and evaluated on the evening images, including both artificial lighting and near-infrared images.
    \item \textbf{External:} The model was trained on images collected by the Top-View and Side-View cameras and evaluated on the external camera images.
\end{itemize}

To study how the training sample size affects model performance in each configuration, the testing set in the cross-validation was fixed to the same 100 images. Then, the training set size was iteratively altered from 16 to 512 images, with the sample size doubled at each step. Each training sample size was repeated 50 times with different random seeds to ensure the robustness of the results. The YOLOv9e, which is the most capable model in the YOLO family to date according to its performance on the COCO dataset, was used as the base model for this study.

\subsection*{Study 2: The Correlation Between Model Complexity and Performance on the Tasks of Localizing Cows}

To investigate whether model performance increases with model complexity, five YOLO-family models were examined in this study. Three of the models were selected from the YOLOv8 family: YOLOv8n, YOLOv8m, and YOLOv8x. All YOLOv8 models share a similar architecture, differing in their depth multiplier, width multiplier, and ratio factor, which collectively determine their parameter counts of 3.2 million (m), 25.9m, and 68.2m, respectively. The depth multiplier determines how many convolutional layers are repeated in a C2F module, the novelty of YOLOv8. The width multiplier and ratio factor collectively specify the channel numbers in the convolutional operations. Correspondingly, YOLOv8n, YOLOv8m, and YOLOv8x are defined by depth multipliers of 0.33, 0.67, and 1.0, respectively. The width multipliers are 0.25, 0.75, and 1.25, while the ratio factors are 2.0, 1.5, and 1.0 \cite{v8yaml}. These variations enable the models to achieve different balances between computational efficiency and accuracy.

The remaining two models were YOLOv9c and YOLOv9e, the latest models in the YOLO family, with parameter counts of 25.6M and 58.2M, respectively. Unlike YOLOv8 models, these models have slightly different backbone architectures. Although the majority of the components between YOLOv9c and YOLOv9e are the same, they primarily differ in their layer counts, module complexities, and depth configurations. YOLOv9c has 618 layers and uses simpler modules, resulting in a more efficient model with lower computational demands. Conversely, YOLOv9e has 1225 layers and employs more advanced modules \cite{v9yaml}. 

All models were trained on 500 images in the five cross-validation configurations: Baseline, Top2Side, Side2Top, Day2Night, and External (\textbf{Figure ~\ref{fig:splits}}). In addition to model performance, computing speed was also evaluated. The training speed was recorded in seconds per 100 epochs on NVIDIA A100 GPUs (NVIDIA, USA), and the inference time was recorded as frames per second (FPS) on both the CPU and GPU (Apple M1 Max chip, Apple Inc., USA). The relationship between model complexity and time consumption was analyzed to provide insights into the trade-off between model performance and computational cost.

\subsection*{Study 3: Assessing the Advantages of Using Fine-Tuned Model Over the Pre-Trained Model as Initial Model Weights}

Most models are released with pre-trained weights obtained from large datasets containing millions of object instances (e.g., COCO \cite{lin2014microsoft} and ImageNet \cite{deng2009imagenet}). The pre-trained models have a general capability in recognizing common objects such as vehicles, animals, and household items. When the model is required to recognize specific objects (i.e., cows in this study), a model trained on a smaller but specific dataset is expected to have better performance. However, such advantages may not necessarily persist as the training sample size increases. Having an equally large number of samples for both the pre-trained and fine-tuned models could diminish the performance gap between the two models. To investigate this hypothesis, this study evaluated the performance of fine-tuned models with two different initial weights. The first initial weight was the default weight from the pre-trained model on the COCO dataset, while the second initial weight was the weight from the fine-tuned model on the opposite view angle. The cross-validation settings are described in \textbf{Table~\ref{tab:fintune_config}}.

\begin{table}[h]
    \caption{Finetuning configurations with different initial weights}
    \centering
    \begin{tabular}{ll}
        \toprule
        \textbf{Finetuning and Prediction Task} & \textbf{Initial Weights} \\
        \midrule
        Top-View Camera & COCO (pre-trained) \\
        & Side-View Camera (fine-tuned) \\
        \midrule
        Side-View Camera & COCO (pre-trained) \\
        & Top-View Camera (fine-tuned) \\
        \midrule
        External Camera & COCO (pre-trained) \\
        & Top-View and Side-View Cameras (fine-tuned) \\
        \bottomrule
    \end{tabular}
    \label{tab:fintune_config}
\end{table}

The backbones of all models (i.e., YOLOv8n, YOLOv8m, YOLOv8x, YOLOv9c, and YOLOv9e) were fine-tuned across different training sample sizes: 16, 32, 64, 128, 256, and 500. The goal was to determine whether the advantage of using the fine-tuned weights persists under the interaction between model complexity and different fine-tuning samples. The performance of the models was evaluated using $\text{mAP@{0.5:0.95}}$.

%% file: _3_results.tex
\section{Results and Discussion}

\subsection*{Public Dataset: COLO}

The COLO dataset contains 1254 images and 11818 cow instances captured from an indoor farm environment. The dataset is organized in YOLO and COCO formats and published on the online platforms GitHub \href{https://github.com/Niche-Squad/COLO/}{(https://github.com/Niche-Squad/COLO/)} and Huggingface \href{https://huggingface.co/datasets/Niche-Squad/COLO}{(https://huggingface.co/datasets/Niche-Squad/COLO)}. The dataset consists of eight configurations (\textbf{Table~\ref{tab:colo_config}}):
\textit{0\_all}, \textit{1\_top}, \textit{2\_side}, \textit{3\_external}, \textit{a1\_t2s}, \textit{a2\_s2t}, \textit{b\_light}, and \textit{c\_external}. The \textit{0\_all} configuration serves as the baseline for this study, featuring non-overlapping training and testing images collected from both the Top-View Camera and Side-View Camera. The \textit{1\_top}, \textit{2\_side}, and \textit{3\_external} configurations contain images from their respective cameras. The \textit{a1\_t2s}, \textit{a2\_s2t}, and \textit{b\_light} configurations include training/testing splits for the Top2Side, Side2Top, and Day2Night scenarios, respectively. The \textit{c\_external} configuration features training images from the Top-View and Side-View Cameras, with testing images from the External Camera. The dataset hosted on GitHub is available as a compressed zip file for public access. In contrast, the dataset on Huggingface requires the Python package "datasets" \cite{datasets} to download. The Huggingface version offers additional functionality to resize the images and annotations to specific resolutions, providing greater flexibility for various applications.

\begin{table}[h]
    \caption{Description of the COLO dataset configurations.}
    \centering
    \begin{tabular}{lll}
        \toprule
        \textbf{Configuration} & \textbf{Training Samples} & \textbf{Testing Samples} \\
        \midrule
        \textit{0\_all} & Top-View + Side-View & Top-View + Side-View \\ [0.5ex]
        \textit{1\_top} & Top-View & Top-View \\[0.5ex]
        \textit{2\_side} & Side-View & Side-View \\[0.5ex]
        \textit{3\_external} & External & External \\[0.5ex]
        \midrule
        \textit{a1\_t2s} & Top-View & Side-View \\[0.5ex]
        \textit{a2\_s2t} & Side-View & Top-View \\[0.5ex]
        \textit{b\_light} & Day & Night \\[0.5ex]
        \textit{c\_external} & Top-View + Side-View & External \\[0.5ex]
        \bottomrule
    \end{tabular}
    \label{tab:colo_config}
\end{table}

\subsection*{Evaluation Metrics}

\begin{figure}[h]
    \centering
    \includegraphics[width=1\textwidth]{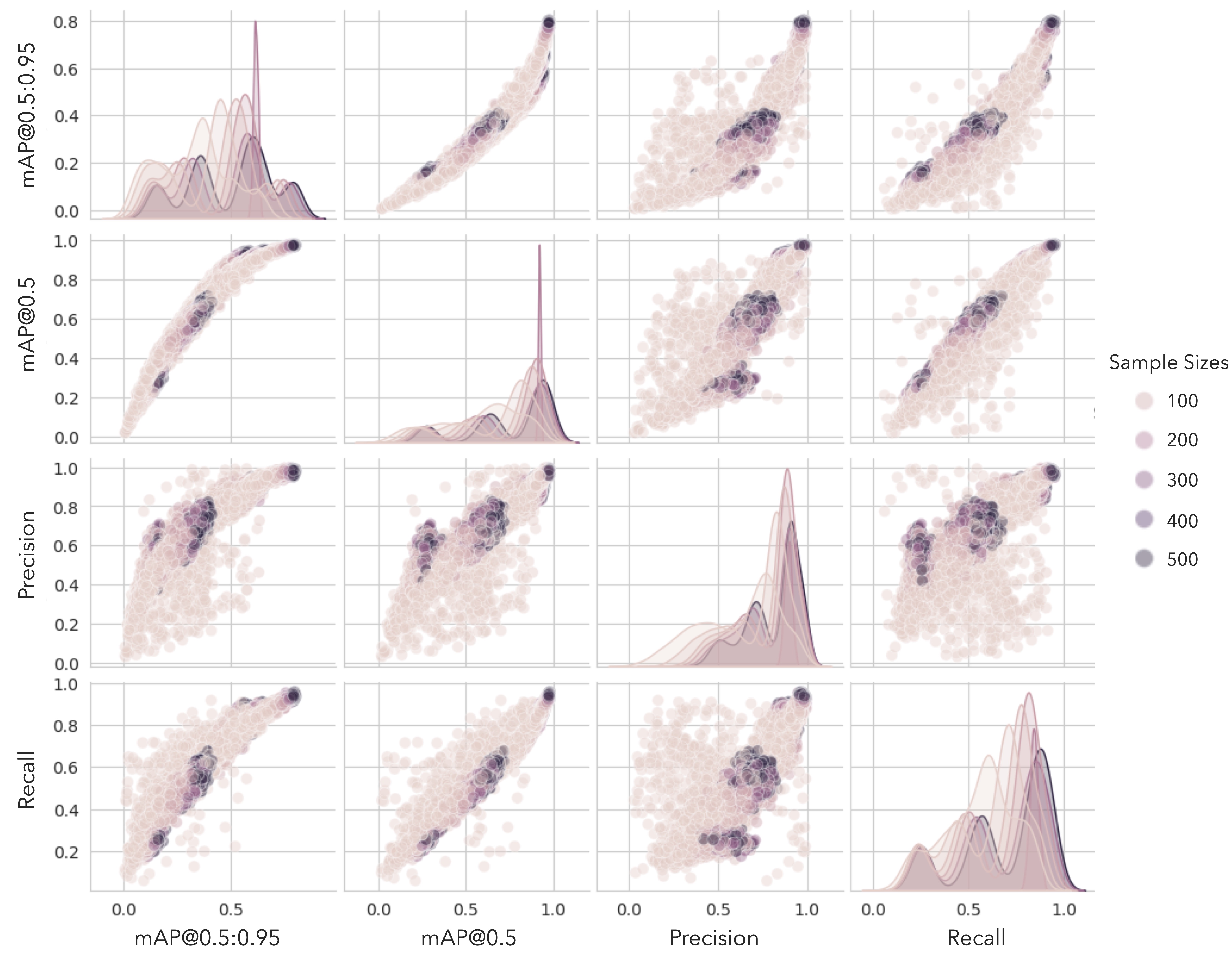}
    \caption{Pairwise scatter plots of the evaluation metrics: $\text{mAP@{0.5:0.95}}$, $\text{mAP@{0.5}}$, precision, and recall. Each point represents a different model configuration, with the color indicating the training sample size.}
    \label{fig:metrics}
\end{figure}

To assess the performance of the YOLO models, we used four key metrics: $\text{mAP@{0.5:0.95}}$, $\text{mAP@{0.5}}$, precision, and recall. These metrics provide a comprehensive understanding of how well the models detect and localize cows in the images from the COLO dataset. A pair-wise comparison of these metrics is presented in \textbf{Figure~\ref{fig:metrics}} to illustrate their interrelationships.

The $\text{mAP@{0.5:0.95}}$ metric is the most stringent, requiring the model to achieve both high positioning accuracy (i.e., high IoU) and high precision across IoU thresholds from 0.5 to 0.95. Because it is less likely to be influenced by high-confidence predictions alone, it serves as a reliable indicator of overall model performance. Achieving an accuracy greater than 0.90 on this metric is generally unrealistic; typically, a value of 0.7 is considered good and is sufficient to yield precision and recall of around 0.9.

In contrast, $\text{mAP@{0.5}}$ is more lenient, requiring high confidence but only moderate IoU. It measures the average precision at an IoU threshold of 0.5. For applications where counting cows is more important than precise positioning, an $\text{mAP@{0.5}}$ value of 0.9 is sufficient. For example, our results showed that the YOLOv8n model, trained on 32 samples, achieved an $\text{mAP@{0.5}}$ of 0.9, making it suitable for such applications.

Precision and recall metrics focus on the accuracy and completeness of the detections. Precision is the ratio of true positive detections to the total number of positive detections (true positives + false positives), measuring how accurate the positive predictions are. Recall is the ratio of true positive detections to the total number of actual positives (true positives + false negatives), measuring the model’s ability to detect all relevant objects. Generally, higher precision is associated with higher recall. However, in some configurations, such as Side2Top and External with smaller sample sizes, models exhibited high recall but low precision. This indicates a tendency to misclassify non-cow objects as cows more frequently than missing actual cows, suggesting a tendency to overestimate rather than underestimate the number of cows in the images.

Our observations emphasize that for applications where counting cows is more critical than precise positioning, achieving a high $\text{mAP@{0.5}}$ is adequate, while the stringent $\text{mAP@{0.5:0.95}}$ metric serves as a comprehensive indicator of overall model performance. These metrics provide insights into both the localization and classification capabilities of the models, helping to identify strengths and weaknesses under different environmental conditions and camera angles.

\subsection*{Study 1: The Changes in Camera View Angles Dramatically Affect Model Performance}

The baseline training configuration showed good generalization capability, with over 90\% of the predictions correctly positioning cows at the 50\% IoU criterion ($\text{mAP@{0.5}}$). The generalization performance can be dissected into changes in view angles (i.e., Top2Side and Side2Top) and lighting conditions (i.e., Day2Night). Changes in lighting conditions did not dramatically affect model performance across all four metrics. However, changing camera views resulted in a performance drop of approximately 30\% and 60\% in $\text{mAP@{0.5}}$ for the Side2Top and Top2Side configurations, respectively. Across all metrics and training sample sizes, the Top2Side configuration consistently showed the worst performance.

\begin{figure}[h]
    \centering
    \includegraphics[width=0.8\textwidth]{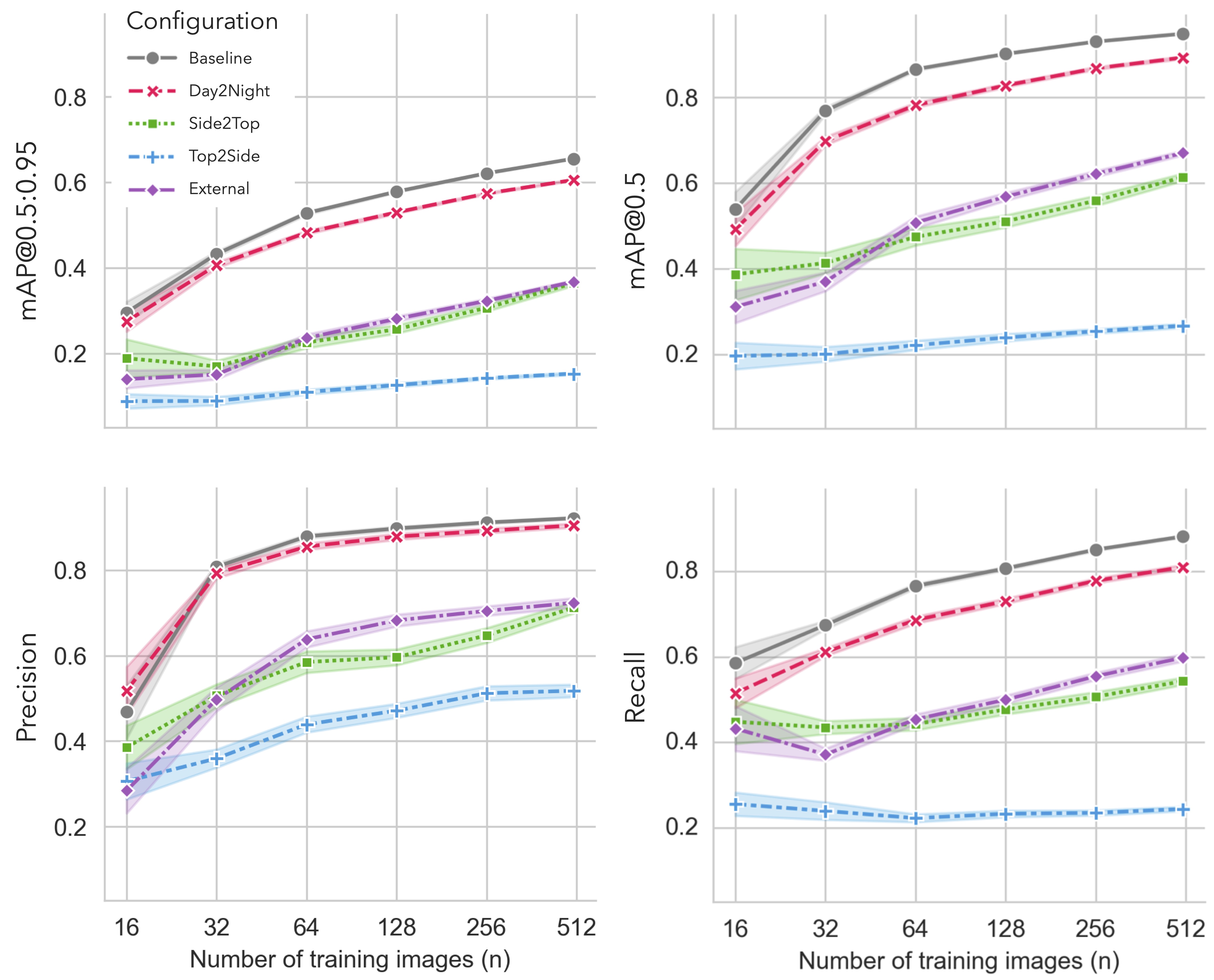}
    \caption{The generalization performance of YOLOv9e across various data configurations and training sample sizes. Sample sizes are depicted on the horizontal axis using a logarithmic scale with a base of 2, and the data configurations are represented by different colors and marker shapes. The upper left and right plots display the metrics $\text{mAP@{0.5:0.95}}$ and $\text{mAP@{0.5}}$, respectively, for different training samples across diverse data configurations. The lower left and right plots depict precision and recall values, also for varying training samples and configurations.}
    \label{fig:schemes}
\end{figure}

From the perspective of precision and recall, changing the camera view from Top2Side resulted in the model missing more than 7 out of every 10 cows, with only 50\% of the detections being correct. For the 'External' configuration, our model identified 6 out of every 10 cows, which is not ideal but also not the worst performance observed. Notably, performance in the Day2Night configuration was close to the baseline in terms of precision, which only considers predictions with high confidence compared to $\text{mAP@{0.5}}$. Hence, by excluding low-confidence predictions, changes in lighting conditions did not affect model performance. Regardless of the configuration and evaluation metrics, model performance always increased as the training sample sizes increased.

This study provides a comparative analysis of the behavior of the model in various data configurations. It is clear that the `Day2Night' configuration shows much better performance relative to the heterogeneous viewpoint-oriented configurations `Top2Side' and `Side2Top'.

Despite the various challenges in adapting models from day to night conditions, the `Day2Night' configuration consistently maintains high precision, closely mirroring the `Baseline' configuration across all training sample sizes. This suggests that changes in lighting have less impact on the model's ability to detect objects compared to changes in viewpoint. This robustness to lighting could be attributed to the inclusion of diverse lighting conditions in the training phase. Specifically, model performance benefited from pixel-wise augmentation techniques such as adjustments to hue, saturation, and value (HSV). These augmentations introduced a variety of color variations to the images, enhancing the model's ability to generalize across different visual conditions. Moreover, these YOLO models benefit from pre-training on the COCO dataset, which is characterized by a wide array of images with varied lighting, aiding their adaptability to shifts in light.

On the other hand, the models perform suboptimally in scenarios involving changes in viewpoint. Each new viewpoint introduces fundamentally different object features that are not replicated through standard data augmentation methods such as lighting or affine transformations. For example, when the camera is placed at a lower angle, cows are more frequently occluded by stalls and fences. These additional objects introduce variations that cannot be mitigated by augmentations in HSV space or image translation. Consequently, `Top2Side' performs the worst, as it is particularly challenging to identify cows from the side. Even for the `External' configuration, the model struggles to generalize well despite being trained on the `Baseline' configuration because the camera angle is changed again in the `External' setup. In summary, camera view angle is crucial for model generalization, with side views being the most challenging.

\subsection*{Study 2: A Higher Model Complexity Does Not Always Lead to Better Performance}

\begin{figure}[h]
\centering
\includegraphics[width=1\textwidth]{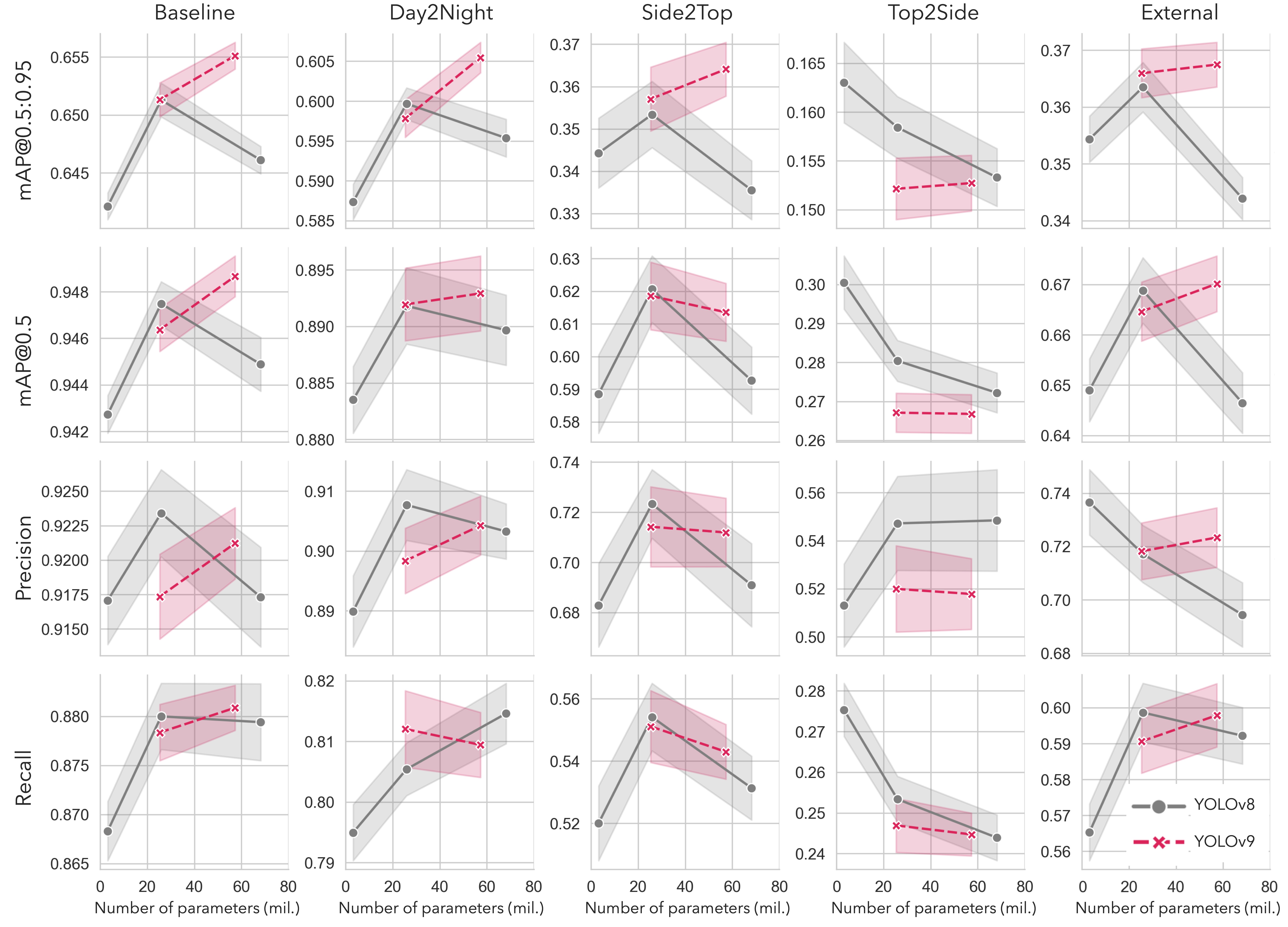}
\caption{The performance of YOLOv8 and YOLOv9 models across various model parameters and data configurations, evaluated using four metrics: $\text{mAP@{0.5:0.95}}$, $\text{mAP@{0.5}}$, precision, and recall. Each column indicates a different data configuration, starting from top left to bottom right: `Baseline', `Day2Night', `Side2Top', `Top2Side', and `External'. The horizontal axis of all plots indicates the number of model parameters.}
\label{fig:models}
\end{figure}

The study found that the training configuration significantly affects the relationship between model complexity and performance. Based on Study 1, predicting images from a side view using a model trained on Top-View camera images is one of the most challenging tasks. In this configuration, increasing model complexity generally resulted in poorer generalization, with simpler models often performing better. However, in other configurations that demonstrated better generalization in Study 1, the peak performance was not always achieved by the most complex model. For example, in the baseline configuration, the YOLOv9e model performed best in terms of $\text{mAP@{0.5:0.95}}$, $\text{mAP@{0.5}}$, and recall, while the YOLOv8m model excelled in precision. Neither of these models had the highest parameter counts compared to YOLOv8x. It is also worth noting that different model architectures showed different performance trends with varying complexities. The YOLOv8-family models tended to perform best with mid-sized models (i.e., YOLOv8m), whereas larger models in the YOLOv9 family usually performed better. Hence, the study concluded that model performance is determined by both the training configuration and the model architecture.

The study results, as shown in \textbf{Figure \ref{fig:models}}, indicate that although both YOLOv8 \cite{ultralyticsYOLOv8} and YOLOv9 \cite{wang2024yolov9} models exhibit an increase in $\text{mAP@{0.5}}$ with more parameters when trained on the COCO dataset \cite{lin2014microsoft}, this does not support a definitive conclusion that more parameters consistently improve model performance. This may be because the prior work's findings were based on the COCO dataset, which includes 80 classes and mainly features standalone images. In contrast, this study uses an indoor farm dataset focused exclusively on a single class: cows. Consequently, the model may not need as many parameters to effectively detect cows. This suggests that researchers should not rely solely on public dataset performance, as model generalization is specific to the task and dataset.

Additionally, this study found that a small model such as YOLOv8n, with only 3.2M parameters, can yield 90\% accuracy with a relatively small size of training samples. This indicates that when one encounters a simple and homogenous task like positioning cows, deploying a small model is optimal in balancing computing time and prediction accuracy. This further underscores the importance of considering the specific characteristics of the task and dataset when choosing a model, rather than defaulting to more complex models under the assumption they will perform better.

Overall, our findings emphasize that higher model complexity does not necessarily lead to better performance. The optimal model configuration depends heavily on the specific task and dataset, highlighting the need for careful model selection tailored to the particular application at hand.

\subsection*{Study 3: The Advantages of Custom Initial Weights are Limited When the Model is Simple}

\begin{figure}[h]
    \centering
    \includegraphics[width=1\textwidth]{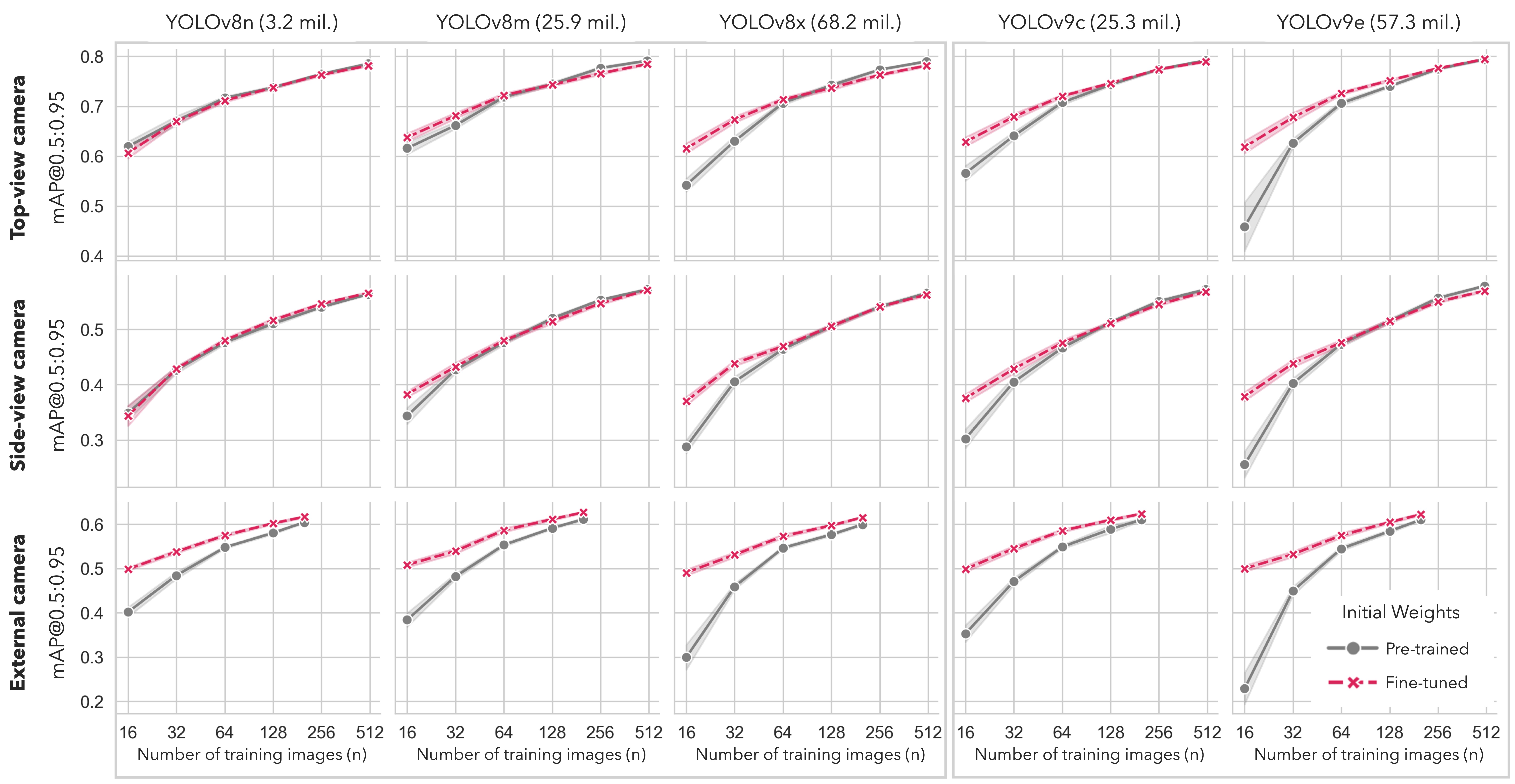}
    \caption{Varied generalization performance in $\text{mAP@{0.5:0.95}}$ with different initial weights. Red lines represent instances where weights were initialized with fine-tuned weights from other data configuration, while grey lines indicate scenarios employing pre-trained weights (i.e., trained with the COCO dataset). The horizontal axis indicates the number of training samples used for the fine-tuning procedure.}
    \label{fig:finetune}
\end{figure}

The results presented in Figure \ref{fig:finetune} indicate that the benefit of using fine-tuned initial weights is minimal for simpler models. Specifically, when employing YOLOv8n, the performance difference between the default and fine-tuned weights was insignificant when fine-tuning data from the Top-View Camera and Side-View Camera. However, as model complexity increased, a greater number of fine-tuning samples were required for the two different initial weights to achieve similar performance. For instance, in the case of YOLOv9e, the performance gap was eliminated when the number of fine-tuning samples reached 128 and 64 for the Top-View Camera and Side-View Camera data sources, respectively. A similar trend was observed with the External camera, where a significant performance gap of more than 25\% in $\text{mAP@{0.5:0.95}}$ was observed for YOLOv9e when the sample size was 16. It is also noted that, although the performance gap was closed to zero for the Top-View Camera and Side-View Camera data sources, the gap was never closed for the External camera.

This study suggests that, for YOLO models with fewer parameters, such as YOLOv8n and YOLOv8m, the choice of weight initialization does not make a significant difference in fine-tuning performance. In contrast, larger models like YOLOv8x, YOLOv9c, and YOLOv9e exhibit improved performance when weights are initialized from a model that has been previously fine-tuned in a similar data configuration, as described in Table \ref{tab:fintune_config}. Therefore, when fine-tuning larger models with a limited dataset, it is beneficial to utilize weights previously fine-tuned on various data configurations. This approach leverages the additional learned features and adaptability from the initial fine-tuning, resulting in better performance even with a small amount of new data. For example, our results showed that YOLOv9e achieved optimal performance with fewer fine-tuning samples when initialized with fine-tuned weights compared to default weights. Conversely, for smaller models, the weight initialization strategy does not significantly impact fine-tuning performance. This is likely due to the lower complexity and fewer parameters of these models, which makes them less dependent on the initial weight configuration to achieve good performance. In practical terms, this means that for simpler models, researchers can save time and computational resources by directly fine-tuning without the need for customized weight initialization.

The analysis of Figure \ref{fig:finetune} also provides insight into performance across homogeneous viewpoint data configurations, specifically `Top-View Camera' and `Side-View Camera'. The data demonstrates that the `Top-View Camera' configuration consistently yields higher mAP values regardless of the training sample size and weight initialization conditions. This implies that the `Side-View Camera' configuration, where both training and test images are captured from the side view, presents a more formidable challenge for cow detection compared to the `Top-View Camera' configuration. The side view poses difficulties due to occlusions by neighboring cows and additional distractions, such as obstacles in aisles and fences. Furthermore, cows located further away in side-view images may not be as visible, complicating feature extraction. In contrast, the `Top-View Camera' configuration benefits from an unobstructed aerial perspective, ensuring that the top view of all cow instances is clearly visible and free from such obstructions. This distinction in visibility between the two configurations contributes to the ease of feature extraction and ultimately, the performance disparity observed.

These findings align with the results from Study 1, which demonstrated that changes in camera view angles dramatically affect model performance. In Study 1, we found that models trained on Top-View datasets struggled the most to detect cows from side-view images, with performance dropping by approximately 60\% in $\text{mAP@{0.5}}$. This significant drop in performance is attributed to the same reasons identified in Study 3: the side view introduces occlusions and distractions that are not present in the top view, making feature extraction more challenging. 

This study highlights that when working with external or unseen datasets, fine-tuning with custom initial weights trained on relevant tasks brings advantages to the detection tasks. On the other hand, simpler models do not benefit much from customized weights, suggesting that it is more efficient to train a simple model with pre-trained weights without relying on prior relevant information, which sometimes requires intensive labor efforts.

\subsection*{Computational Resource Requirements}

\begin{figure}[h]
    \centering
    \includegraphics[width=.8\textwidth]{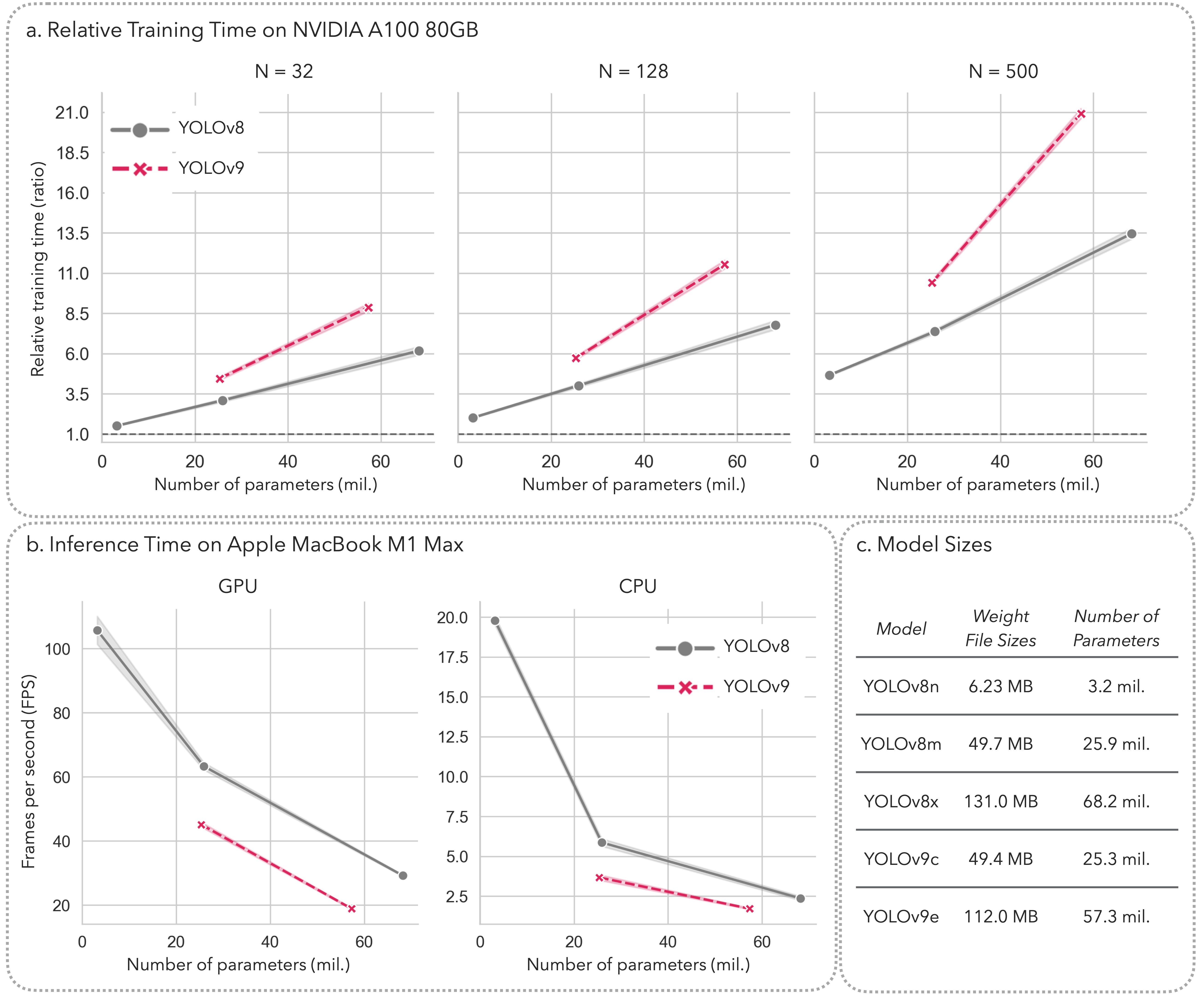}
    \caption{Comparative evaluation of computational resource requirements. (a) Training time (expressed as a multiple of baseline time) versus number of parameters for YOLOv8 and YOLOv9 models, presented for training sample sizes of 32 (left), 128 (middle), and 500 (right). (b) Inference frequency versus number of parameters for YOLOv8 and YOLOv9 models on GPU (left) and CPU (right). (c) A table displaying the weight sizes and parameter counts of various YOLOv8 and YOLOv9 models.}
    \label{fig:resources}
\end{figure}

The evaluation of computational resource requirements is crucial for understanding the feasibility of deploying YOLO models in real-world applications, especially in environments with limited computational resources. This section compares training time (Figure \ref{fig:resources}a), inference time (Figure \ref{fig:resources}b), and model weight sizes (Figure \ref{fig:resources}c) for various YOLO models.

The training time for each model was measured and expressed as a multiple of the baseline training time, which is the time required to train the YOLOv8n model with 32 samples. The results indicated that using the largest model, YOLOv8x, which has 20 times more parameters, increased training time by 4 to 6 times, depending on the training sample size. Additionally, the YOLOv9 models generally required more training time and had slower inference frames per second (FPS) compared to the YOLOv8 models. The gap in training time expanded as the number of training samples increased.

Inference time was measured as the average FPS in a batch of 64 images. Running the models on a CPU with the smallest model (YOLOv8n) was slower than running the largest model (YOLOv8x) on a GPU. For example, the FPS for the small, YOLOv8n, on a CPU was 19.77, while the FPS for YOLOv8x on a GPU was 29.21. High FPS models are essential for real-time inference, which usually requires a model with an FPS higher than 30. The results indicate that implementing YOLO models on a CPU may not meet real-time requirements, especially for larger models.

Lastly, model weight sizes were also considered, impacting memory requirements and deployment feasibility, especially in edge computing environments. The weight sizes and parameter counts of various YOLO models are displayed in Figure \ref{fig:resources}c.

In conclusion, this evaluation highlights the trade-offs between model complexity and computational efficiency. The larger YOLO models, while offering potentially better performance, require significantly more computational resources. This analysis helps researchers and practitioners select the appropriate model based on the available computational resources and the specific requirements of their application.

\section{Conclusion}

This study examined the impact of various training configurations and model complexities on the performance of YOLOv8 and YOLOv9 models for cow detection in indoor farm environments. Our results indicate that model performance is highly dependent on camera viewpoints, with side views presenting the greatest challenges. Additionally, fine-tuning models with weights from similar datasets substantially enhances performance, particularly for complex models in scenarios with limited data. We also introduce a public cow localization dataset, 'COLO', to support the research community.

The findings indicate that while increasing model complexity can improve performance, this is not always the case, especially in challenging configurations like 'Top2Side', which predict images from a side view using a model trained on top-view images. Models trained on a single viewpoint exhibit limited generalization, underscoring the importance of incorporating diverse and consistent camera angles in the training data.

Despite the promising results, this study has certain limitations. The models' performance was evaluated under specific indoor farm conditions, which may not generalize to all livestock environments. Moreover, the reliance on pre-defined configurations may limit the applicability of our findings to more dynamic settings.

Future work should explore adaptive methods for enhancing model generalization across varied viewpoints and environmental conditions. Additionally, investigating the integration of advanced data augmentation techniques and more diverse datasets could further improve detection accuracy and robustness.

In conclusion, this study offers practical insights into reproducing model performance in new environmental settings and provides the public 'COLO' dataset to facilitate further research and advancements in the field.

%% file: _9_appendix.tex
\section*{Appendix}

\subsection*{Hyperparameters in Ultralytics library}

The table below show the hyperparameters used in the Ultralytics library for training the models in this study.

\begin{table}[h]
    \caption{Hyperparameters for the training procedure}
    \centering
    \begin{tabular}{lp{0.5\textwidth}l}
        \toprule
        Hyperparameters & Description & Value \\
        \midrule
        epochs & Number of training epochs & 100 \\
        batch & Number of images in each batch & 16 \\
        optimizer & Optimizer used for training & auto \\
        \text{hsv\_h} & Altering the hue value of the image & 0.015 \\
        \text{hsv\_s} & Altering the saturation of the image by a fraction & 0.7 \\
        \text{hsv\_v} & Altering the brightness of the image by a fraction & 0.4 \\
        translate & Randomly translating the image by a fraction of the image size & 0.1 \\
        scale & Randomly scaling the image by a fraction of the image size & 0.5 \\
        fliplr & Randomly flipping the image horizontally with the given probability & 0.5 \\
        mosaic & Combining four images into one mosaic image with the given probability & 1.0 \\
        mixup & Randomly mixing up the object instances across multiple images with the given probability & 0.15 \\
        \text{copy\_paste} & Randomly copying and pasting the object instances across multiple images with the given probability & 0.3 \\
        \bottomrule
    \end{tabular}
    \label{tab:hyperparameters}
\end{table}

%% file: main.bbl
\begin{thebibliography}{54}
\providecommand{\natexlab}[1]{#1}
\providecommand{\url}[1]{\texttt{#1}}
\expandafter\ifx\csname urlstyle\endcsname\relax
  \providecommand{\doi}[1]{doi: #1}\else
  \providecommand{\doi}{doi: \begingroup \urlstyle{rm}\Url}\fi

\bibitem[COL(2023)]{COLODataset2023}
Colo: A large-scale dataset for conversational question answering over
  knowledge graphs.
\newblock \url{https://huggingface.co/datasets/Niche-Squad/COLO}, 2023.
\newblock Accessed: 2024-04-09.

\bibitem[v8y(2023)]{v8yaml}
Ultralytics v8 models.
\newblock
  \url{https://github.com/ultralytics/ultralytics/blob/main/ultralytics/cfg/models/v8/yolov8.yaml},
  2023.

\bibitem[v9y(2023)]{v9yaml}
Ultralytics v9 models.
\newblock
  \url{https://github.com/ultralytics/ultralytics/tree/main/ultralytics/cfg/models/v9},
  2023.

\bibitem[Deng et~al.(2009)Deng, Dong, Socher, Li, Li, and
  Fei-Fei]{deng2009imagenet}
Jia Deng, Wei Dong, Richard Socher, Li-Jia Li, Kai Li, and Li~Fei-Fei.
\newblock Imagenet: A large-scale hierarchical image database.
\newblock In \emph{2009 IEEE conference on computer vision and pattern
  recognition}, pages 248--255. Ieee, 2009.

\bibitem[Duan et~al.(2019)Duan, Bai, Xie, Qi, Huang, and
  Tian]{duan2019centernet}
K.~Duan, S.~Bai, L.~Xie, H.~Qi, Q.~Huang, and Q.~Tian.
\newblock Centernet: Keypoint triplets for object detection.
\newblock In \emph{Proceedings of the IEEE/CVF International Conference on
  Computer Vision}, pages 6569--6578, Seoul, Republic of Korea, 27 October--2
  November 2019.

\bibitem[Elfwing et~al.(2018)Elfwing, Uchibe, and Doya]{elfwing2018sigmoid}
Stefan Elfwing, Eiji Uchibe, and Kenji Doya.
\newblock Sigmoid-weighted linear units for neural network function
  approximation in reinforcement learning.
\newblock \emph{Neural networks}, 107:\penalty0 3--11, 2018.

\bibitem[Fernandes et~al.(2020)Fernandes, D{\'o}rea, and
  Rosa]{fernandes2020image}
Arthur Francisco~Ara{\'u}jo Fernandes, Jo{\~a}o Ricardo~Rebou{\c{c}}as
  D{\'o}rea, and Guilherme Jord{\~a}o de~Magalh{\~a}es Rosa.
\newblock Image analysis and computer vision applications in animal sciences:
  an overview.
\newblock \emph{Frontiers in Veterinary Science}, 7:\penalty0 551269, 2020.

\bibitem[Girshick(2015)]{girshick2015fast}
Ross Girshick.
\newblock Fast r-cnn.
\newblock In \emph{Proceedings of the IEEE international conference on computer
  vision}, pages 1440--1448, 2015.

\bibitem[Greif(2024)]{greif_dgreifring_2024}
Dusty Greif.
\newblock dgreif/ring, April 2024.
\newblock URL \url{https://github.com/dgreif/ring}.
\newblock original-date: 2018-10-12T22:53:01Z.

\bibitem[Guirguis et~al.(2022)Guirguis, Hendawy, Eskandar, Abdelsamad, Kayser,
  and Beyerer]{guirguis2022cfa}
Karim Guirguis, Ahmed Hendawy, George Eskandar, Mohamed Abdelsamad, Matthias
  Kayser, and J{\"u}rgen Beyerer.
\newblock Cfa: Constraint-based finetuning approach for generalized few-shot
  object detection.
\newblock In \emph{Proceedings of the IEEE/CVF conference on computer vision
  and pattern recognition}, pages 4039--4049, 2022.

\bibitem[Gupta et~al.(2023)Gupta, Gill, Gulia, and Chatterjee]{gupta2023novel}
Chhaya Gupta, Nasib~Singh Gill, Preeti Gulia, and Jyotir~Moy Chatterjee.
\newblock A novel finetuned yolov6 transfer learning model for real-time object
  detection.
\newblock \emph{Journal of Real-Time Image Processing}, 20\penalty0
  (3):\penalty0 42, 2023.

\bibitem[Han et~al.(2021)Han, Zhang, Ding, Gu, Liu, Huo, Qiu, Yao, Zhang,
  Zhang, et~al.]{han2021pre}
Xu~Han, Zhengyan Zhang, Ning Ding, Yuxian Gu, Xiao Liu, Yuqi Huo, Jiezhong Qiu,
  Yuan Yao, Ao~Zhang, Liang Zhang, et~al.
\newblock Pre-trained models: Past, present and future.
\newblock \emph{AI Open}, 2:\penalty0 225--250, 2021.

\bibitem[Hao et~al.(2023)Hao, Ren, Han, Zhang, Li, and Liu]{hao2023cattle}
Wangli Hao, Chao Ren, Meng Han, Li~Zhang, Fuzhong Li, and Zhenyu Liu.
\newblock Cattle body detection based on yolov5-ema for precision livestock
  farming.
\newblock \emph{Animals}, 13\penalty0 (22):\penalty0 3535, 2023.

\bibitem[Hariharan et~al.(2015)Hariharan, Arbel{\'a}ez, Girshick, and
  Malik]{hariharan2015hypercolumns}
Bharath Hariharan, Pablo Arbel{\'a}ez, Ross Girshick, and Jitendra Malik.
\newblock Hypercolumns for object segmentation and fine-grained localization.
\newblock In \emph{Proceedings of the IEEE conference on computer vision and
  pattern recognition}, pages 447--456, 2015.

\bibitem[He et~al.(2016)He, Zhang, Ren, and Sun]{he2016deep}
Kaiming He, Xiangyu Zhang, Shaoqing Ren, and Jian Sun.
\newblock Deep residual learning for image recognition.
\newblock In \emph{Proceedings of the IEEE conference on computer vision and
  pattern recognition}, pages 770--778, 2016.

\bibitem[He et~al.(2017)He, Gkioxari, Doll{\'a}r, and Girshick]{he2017mask}
Kaiming He, Georgia Gkioxari, Piotr Doll{\'a}r, and Ross Girshick.
\newblock Mask r-cnn.
\newblock In \emph{Proceedings of the IEEE international conference on computer
  vision}, pages 2961--2969, 2017.

\bibitem[Hu et~al.(2021)Hu, Chu, Pei, Liu, and Bian]{hu2021model}
Xia Hu, Lingyang Chu, Jian Pei, Weiqing Liu, and Jiang Bian.
\newblock Model complexity of deep learning: A survey.
\newblock \emph{Knowledge and Information Systems}, 63:\penalty0 2585--2619,
  2021.

\bibitem[Huang et~al.(2017)Huang, Liu, Van Der~Maaten, and
  Weinberger]{huang2017densely}
Gao Huang, Zhuang Liu, Laurens Van Der~Maaten, and Kilian~Q Weinberger.
\newblock Densely connected convolutional networks.
\newblock In \emph{Proceedings of the IEEE conference on computer vision and
  pattern recognition}, pages 4700--4708, 2017.

\bibitem[Huang et~al.(2019)Huang, Huang, Gong, Huang, and Wang]{huang2019mask}
Zhaojin Huang, Lichao Huang, Yongchao Gong, Chang Huang, and Xinggang Wang.
\newblock Mask scoring r-cnn.
\newblock In \emph{Proceedings of the IEEE/CVF conference on computer vision
  and pattern recognition}, pages 6409--6418, 2019.

\bibitem[Jocher(2020)]{Jocher2020YOLOv5}
Glenn Jocher.
\newblock {YOLOv5 by Ultralytics}.
\newblock \url{https://github.com/ultralytics/yolov5}, 2020.
\newblock Accessed on: 28 February 2023.

\bibitem[Justus et~al.(2018)Justus, Brennan, Bonner, and
  McGough]{justus2018predicting}
Daniel Justus, John Brennan, Stephen Bonner, and Andrew~Stephen McGough.
\newblock Predicting the computational cost of deep learning models.
\newblock In \emph{2018 IEEE international conference on big data (Big Data)},
  pages 3873--3882. IEEE, 2018.

\bibitem[Karen(2014)]{karen2014very}
Simonyan Karen.
\newblock Very deep convolutional networks for large-scale image recognition.
\newblock \emph{arXiv preprint arXiv: 1409.1556}, 2014.

\bibitem[Krizhevsky et~al.(2017)Krizhevsky, Sutskever, and
  Hinton]{krizhevsky2017imagenet}
Alex Krizhevsky, Ilya Sutskever, and Geoffrey~E Hinton.
\newblock Imagenet classification with deep convolutional neural networks.
\newblock \emph{Communications of the ACM}, 60\penalty0 (6):\penalty0 84--90,
  2017.

\bibitem[Law and Deng(2018)]{law2018cornernet}
H.~Law and J.~Deng.
\newblock Cornernet: Detecting objects as paired keypoints.
\newblock In \emph{Proceedings of the European Conference on Computer Vision
  (ECCV)}, pages 734--750, Munich, Germany, 8--14 September 2018.

\bibitem[Lhoest et~al.(2021)Lhoest, Villanova~del Moral, Jernite, Thakur, von
  Platen, Patil, Chaumond, Drame, Plu, Tunstall, Davison, {\v{C}}uklina,
  Brandeis, Le~Scao, Schmid, Gugger, Delangue, Matussi{\`{e}}re, Debut,
  Ben-Ami, Filippova, d'Hoffschmidt, G{\'{e}}rard, Lane, Ansell, Buitinck,
  Esposito, Raison, Klein, Nguyen, Mikami, Sanh, Chaudhary, Patry, Chang,
  Froment, Buhmann, Malartic, Winschel, Watson, Pradeep, Chhablani, Rohrbach,
  Jenny, Bolton, Phang, Löw, Rush, and Wolf]{datasets}
Quentin Lhoest, Albert Villanova~del Moral, Yacine Jernite, Abhishek Thakur,
  Patrick von Platen, Suraj Patil, Julien Chaumond, Mariama Drame, Julien Plu,
  Lewis Tunstall, Joe Davison, Marija {\v{C}}uklina, Simon Brandeis, Teven
  Le~Scao, Philipp Schmid, Sylvain Gugger, Cl{\'{e}}ment Delangue, Th{\'{e}}o
  Matussi{\`{e}}re, Lysandre Debut, Idan Ben-Ami, Olga Filippova, Martin
  d'Hoffschmidt, Sebastien G{\'{e}}rard, Brendan Lane, Leo Ansell, Lars
  Buitinck, Damien Esposito, Mathis Raison, Jacob Klein, Thibault Nguyen,
  Tomoki Mikami, Victor Sanh, Vishrav Chaudhary, Nicolas Patry, Wilson~Y.
  Chang, Julien Froment, Jonas Buhmann, Quentin Malartic, Victor Winschel,
  Charlie Watson, Rajarshi Pradeep, Gunjan Chhablani, Manuela Rohrbach, Maxim
  Jenny, John Bolton, Jason Phang, Theo Löw, Alexander Rush, and Thomas Wolf.
\newblock Datasets: A community library for natural language processing.
\newblock \url{https://github.com/huggingface/datasets}, 2021.

\bibitem[Li et~al.(2021)Li, Huang, Chen, Chesser~Jr, Purswell, Linhoss, and
  Zhao]{li2021practices}
Guoming Li, Yanbo Huang, Zhiqian Chen, Gary~D Chesser~Jr, Joseph~L Purswell,
  John Linhoss, and Yang Zhao.
\newblock Practices and applications of convolutional neural network-based
  computer vision systems in animal farming: A review.
\newblock \emph{Sensors}, 21\penalty0 (4):\penalty0 1492, 2021.

\bibitem[Lin et~al.(2014)Lin, Maire, Belongie, Hays, Perona, Ramanan,
  Doll{\'a}r, and Zitnick]{lin2014microsoft}
Tsung-Yi Lin, Michael Maire, Serge Belongie, James Hays, Pietro Perona, Deva
  Ramanan, Piotr Doll{\'a}r, and C~Lawrence Zitnick.
\newblock Microsoft coco: Common objects in context.
\newblock In \emph{Computer Vision--ECCV 2014: 13th European Conference,
  Zurich, Switzerland, September 6-12, 2014, Proceedings, Part V 13}, pages
  740--755. Springer, 2014.

\bibitem[Liu et~al.(2016)Liu, Anguelov, Erhan, Szegedy, Reed, Fu, and
  Berg]{liu2016ssd}
Wei Liu, Dragomir Anguelov, Dumitru Erhan, Christian Szegedy, Scott Reed,
  Cheng-Yang Fu, and Alexander~C Berg.
\newblock Ssd: Single shot multibox detector.
\newblock In \emph{Computer Vision--ECCV 2016: 14th European Conference,
  Amsterdam, The Netherlands, October 11--14, 2016, Proceedings, Part I 14},
  pages 21--37. Springer, 2016.

\bibitem[{Massachusetts Institute of Technology}(2023)]{labelme2023}
{Massachusetts Institute of Technology}.
\newblock Labelme: Image annotation tool.
\newblock \url{http://labelme.csail.mit.edu/Release3.0/}, 2023.

\bibitem[Morrone et~al.(2022)Morrone, Dimauro, Gambella, and
  Cappai]{morrone2022industry}
Sarah Morrone, Corrado Dimauro, Filippo Gambella, and Maria~Grazia Cappai.
\newblock Industry 4.0 and precision livestock farming (plf): an up to date
  overview across animal productions.
\newblock \emph{Sensors}, 22\penalty0 (12):\penalty0 4319, 2022.

\bibitem[Nasirahmadi et~al.(2019)Nasirahmadi, Sturm, Edwards, Jeppsson, Olsson,
  M{\"u}ller, and Hensel]{nasirahmadi2019deep}
Abozar Nasirahmadi, Barbara Sturm, Sandra Edwards, Knut-H{\aa}kan Jeppsson,
  Anne-Charlotte Olsson, Simone M{\"u}ller, and Oliver Hensel.
\newblock Deep learning and machine vision approaches for posture detection of
  individual pigs.
\newblock \emph{Sensors}, 19\penalty0 (17):\penalty0 3738, 2019.

\bibitem[Ninja(2024)]{visualization-tools-for-opencows2020-dataset}
Dataset Ninja.
\newblock Visualization tools for opencow2020 dataset.
\newblock \url{ https://datasetninja.com/opencows2020 }, may 2024.
\newblock URL \url{https://datasetninja.com/opencows2020}.
\newblock visited on 2024-05-21.

\bibitem[Noe et~al.(2022)Noe, Zin, Tin, and Kobayashi]{noe2022automatic}
Su~Myat Noe, Thi~Thi Zin, Pyke Tin, and Ikuo Kobayashi.
\newblock Automatic detection and tracking of mounting behavior in cattle using
  a deep learning-based instance segmentation model.
\newblock \emph{Int. J. Innov. Comput. Inf. Control}, 18\penalty0 (1):\penalty0
  211--220, 2022.

\bibitem[Ong et~al.(2023)Ong, Retta, Srinivasan, Tan, and
  Liu]{ong2023cattleeyeview}
Kian~Eng Ong, Sivaji Retta, Ramarajulu Srinivasan, Shawn Tan, and Jun Liu.
\newblock Cattleeyeview: A multi-task top-down view cattle dataset for smarter
  precision livestock farming.
\newblock In \emph{2023 IEEE International Conference on Visual Communications
  and Image Processing (VCIP)}, pages 1--5. IEEE, 2023.

\bibitem[{OpenCV}(2023)]{cvat2023}
{OpenCV}.
\newblock Cvat: Computer vision annotation tool.
\newblock \url{https://www.cvat.ai/}, 2023.

\bibitem[Redmon et~al.(2016)Redmon, Divvala, Girshick, and
  Farhadi]{redmon2016you}
Joseph Redmon, Santosh Divvala, Ross Girshick, and Ali Farhadi.
\newblock You only look once: Unified, real-time object detection.
\newblock In \emph{Proceedings of the IEEE conference on computer vision and
  pattern recognition}, pages 779--788, 2016.

\bibitem[{Roboflow}(2023)]{roboflow2023}
{Roboflow}.
\newblock Roboflow: Organize, annotate, and improve machine learning datasets.
\newblock \url{https://roboflow.com/}, 2023.

\bibitem[Siddique et~al.(2021)Siddique, Paheding, Elkin, and
  Devabhaktuni]{siddique2021u}
Nahian Siddique, Sidike Paheding, Colin~P Elkin, and Vijay Devabhaktuni.
\newblock U-net and its variants for medical image segmentation: A review of
  theory and applications.
\newblock \emph{Ieee Access}, 9:\penalty0 82031--82057, 2021.

\bibitem[Simonyan and Zisserman(2014)]{simonyan2014very}
Karen Simonyan and Andrew Zisserman.
\newblock Very deep convolutional networks for large-scale image recognition.
\newblock \emph{arXiv preprint arXiv:1409.1556}, 2014.

\bibitem[Szegedy et~al.(2015)Szegedy, Liu, Jia, Sermanet, Reed, Anguelov,
  Erhan, Vanhoucke, and Rabinovich]{szegedy2015going}
Christian Szegedy, Wei Liu, Yangqing Jia, Pierre Sermanet, Scott Reed, Dragomir
  Anguelov, Dumitru Erhan, Vincent Vanhoucke, and Andrew Rabinovich.
\newblock Going deeper with convolutions.
\newblock In \emph{Proceedings of the IEEE conference on computer vision and
  pattern recognition}, pages 1--9, 2015.

\bibitem[T.~Psota et~al.(2020)T.~Psota, Schmidt, Mote, and
  C.~P{\'e}rez]{t2020long}
Eric T.~Psota, Ty~Schmidt, Benny Mote, and Lance C.~P{\'e}rez.
\newblock Long-term tracking of group-housed livestock using keypoint detection
  and map estimation for individual animal identification.
\newblock \emph{Sensors}, 20\penalty0 (13):\penalty0 3670, 2020.

\bibitem[Targ et~al.(2016)Targ, Almeida, and Lyman]{targ2016resnet}
Sasha Targ, Diogo Almeida, and Kevin Lyman.
\newblock Resnet in resnet: Generalizing residual architectures.
\newblock \emph{arXiv preprint arXiv:1603.08029}, 2016.

\bibitem[Tian et~al.(2019)Tian, Shen, Chen, and He]{tian2019fcos}
Z.~Tian, C.~Shen, H.~Chen, and T.~He.
\newblock Fcos: Fully convolutional one-stage object detection.
\newblock In \emph{Proceedings of the IEEE/CVF International Conference on
  Computer Vision}, pages 9627--9636, Seoul, Republic of Korea, 27 October--2
  November 2019.

\bibitem[Tishby and Zaslavsky(2015)]{tishby2015deep}
Naftali Tishby and Noga Zaslavsky.
\newblock Deep learning and the information bottleneck principle.
\newblock In \emph{2015 ieee information theory workshop (itw)}, pages 1--5.
  IEEE, 2015.

\bibitem[Tishby et~al.(2000)Tishby, Pereira, and Bialek]{tishby2000information}
Naftali Tishby, Fernando~C Pereira, and William Bialek.
\newblock The information bottleneck method.
\newblock \emph{arXiv preprint physics/0004057}, 2000.

\bibitem[Tu et~al.(2021)Tu, Yuan, Liang, Wang, and Wan]{tu2021automatic}
Shuqin Tu, Weijun Yuan, Yun Liang, Fan Wang, and Hua Wan.
\newblock Automatic detection and segmentation for group-housed pigs based on
  pigms r-cnn.
\newblock \emph{Sensors}, 21\penalty0 (9):\penalty0 3251, 2021.

\bibitem[Ultralytics()]{ultralytics}
Ultralytics.
\newblock Ultralytics models documentation.
\newblock \url{https://docs.ultralytics.com/models/}.
\newblock Accessed: 2024-05-21.

\bibitem[Ultralytics(2023)]{ultralytics2023datasets}
Ultralytics.
\newblock Ultralytics datasets documentation.
\newblock \url{https://docs.ultralytics.com/datasets/detect/}, 2023.

\bibitem[Ultralytics(Januray 2023)]{ultralyticsYOLOv8}
Ultralytics.
\newblock {Y}{O}{L}{O}v8 --- docs.ultralytics.com.
\newblock \url{https://docs.ultralytics.com/models/yolov8/#overview}, Januray
  2023.

\bibitem[Viola and Jones(2001)]{viola2001rapid}
Paul Viola and Michael Jones.
\newblock Rapid object detection using a boosted cascade of simple features.
\newblock In \emph{Proceedings of the 2001 IEEE computer society conference on
  computer vision and pattern recognition. CVPR 2001}, volume~1, pages I--I.
  Ieee, 2001.

\bibitem[Wang and Liao(2024)]{wang2024yolov9}
Chien-Yao Wang and Hong-Yuan~Mark Liao.
\newblock {YOLOv9}: Learning what you want to learn using programmable gradient
  information.
\newblock 2024.

\bibitem[Xu and Wu(2020)]{xu2020improved}
Danqing Xu and Yiquan Wu.
\newblock Improved yolo-v3 with densenet for multi-scale remote sensing target
  detection.
\newblock \emph{Sensors}, 20\penalty0 (15):\penalty0 4276, 2020.

\bibitem[Yu et~al.(2022)Yu, Liu, Yu, Wang, Song, Yan, Li, Wang, and
  Tian]{yu2022automatic}
Zhenwei Yu, Yuehua Liu, Sufang Yu, Ruixue Wang, Zhanhua Song, Yinfa Yan, Fade
  Li, Zhonghua Wang, and Fuyang Tian.
\newblock Automatic detection method of dairy cow feeding behaviour based on
  yolo improved model and edge computing.
\newblock \emph{Sensors}, 22\penalty0 (9):\penalty0 3271, 2022.

\bibitem[Zin et~al.(2020)Zin, Pwint, Seint, Thant, Misawa, Sumi, and
  Yoshida]{zin_automatic_2020}
Thi~Thi Zin, Moe~Zet Pwint, Pann~Thinzar Seint, Shin Thant, Shuhei Misawa,
  Kosuke Sumi, and Kyohiro Yoshida.
\newblock Automatic {Cow} {Location} {Tracking} {System} {Using} {Ear} {Tag}
  {Visual} {Analysis}.
\newblock \emph{Sensors}, 20\penalty0 (12):\penalty0 3564, January 2020.
\newblock ISSN 1424-8220.
\newblock \doi{10.3390/s20123564}.
\newblock URL \url{https://www.mdpi.com/1424-8220/20/12/3564}.
\newblock Number: 12 Publisher: Multidisciplinary Digital Publishing Institute.

\end{thebibliography}
